# Performance Prediction of Data-Driven Knowledge summarization of High Entropy Alloys (HEAs) literature implementing Natural Language Processing algorithms

Akshansh Mishra[1], Vijaykumar S. Jatti[2], Vaishnavi More[3], Anish Dasgupta[4], Devarrishi Dixit[5] and Eyob Messele Sefene[6]

[1]School of Industrial and Information Engineering, Politecnico Di Milano, Milan, Italy

[2]Symbiosis Institute of Technology, Symbiosis International (Deemed) University, Pune 412115, India

[3]Software Developer, Turabit LLC, Ahmedabad, India

[4]Data Science Engineer, Volvo Groups, Bengaluru, India

[5]Department of Materials Science Engineering, Christian Albrechts University zu Kiel 24143, Germany

[6]Department of Mechanical Engineering, National Taiwan University of Science and Technology, Taiwan

**Abstract:** The ability to interpret spoken language is connected to natural language processing. It involves teaching the AI how words relate to one another, how they are meant to be used, and in what settings. The goal of natural language processing (NLP) is to get a machine intelligence to process words the same way a human brain does. This enables machine intelligence to interpret, arrange, and comprehend textual data by processing the natural language. The technology can comprehend what is communicated, whether it be through speech or writing because AI pro-cesses language more quickly than humans can. In the present study, five NLP algorithms, namely, Geneism, Sumy, Luhn, Latent Semantic Analysis (LSA), and Kull-back-Liebler (KL) al-gorithm, are implemented for the first time for the knowledge summarization purpose of the High Entropy Alloys (HEAs). The performance prediction of these algorithms is made by using the BLEU score and ROUGE score. The results showed that the Luhn algorithm has the highest accuracy score for the knowledge summarization tasks compared to the other used algorithms.

## 1. Introduction

Natural language processing is the ability of a computer program to interpret natural language, or a system of communication, in both its spoken and written forms (NLP). It falls under artificial intelligence (AI) and Linguistics has origins in NLP, which has been around for over 50 years. It is useful in a variety of fields, including as search engines, corporate intelligence, and medical research. [1-3]. NLP has made it possible for computers to understand natural language in the same way that humans do. Whether the speech is written or spoken, natural language processing uses artificial intelligence to gather direct experience, analyse it, and organize it logically in a way that a computer can understand. Similar to how individuals have multiple sensors like ears to listen and eyes to see, computers have reading programs and speakers to gather sounds. Systems have a script to process their various inputs, just as individuals have a mind to do so. The input is eventually translated into computer-readable code during processing [4-5]. Businesses need a way to effectively process the vast amounts of unorganized, text-heavy data they use. Until recently, businesses could not efficiently analyse the natural human language that made up a large portion of the information generated online and kept in databases. Natural language processing comes in handy in this situation.

Consider the following two sentences: "Cloud computing coverage should be included of every service-level agreement" and "A solid SLA guarantees an easier night's sleep-even in the cloud" to see the benefit of NLP. If a user uses NLP to conduct a search, the software will understand that cloud computing is a concept, that cloud is an acronym for cloud computing, and that SLA is an abbreviation for service-level agreement. For various purposes, including data gathering, hypothesis building, and pattern detection within and across domains, NLP can be employed for materials, as stated by Olivetti et al. [6] and illustrated with instances. Tshitoyan et al. [7] employed unsuperived method for materials. The earlier articles contained significant implicit knowledge about upcoming findings. This further demonstrates that earlier articles include a substantial number of implicit knowledge about forthcoming findings. Their results, which hint at a complete technique for mining scientific literature, emphasized the possibility of jointly retrieving information and connections from a large amount of scientific literature. Mohammadi et al. [8] used 1120 articles published between 1974 to 2007 and evaluated the transdisciplinary tendencies of Iranian research in NST. Using text mining tools, it was discovered that the fundamental ideas of the Iranian literature included in NST were contained in 96 phrases. By using multidimensional scaling and basing it on the co-occurrence of the essential phrases in the academic articles, the scientific backbone of the Iranian NST was then reconstructed. The findings demonstrated that the multidisciplinary structure of the NST domain in Iranian papers includes pure physics, analytical chemistry, material science and engineering, biochemistry, chemical physics, physical science, and new developing themes. The analysis indicates that whereas most RMC literature has been ad-dressed to concrete technology and material science, very few studies have concentrated on RMC dispatching [9].

Finding the root reasons and pertinent elements is necessary to learn from past mistakes. A valuable learning resource is the enormous corpus of textual data on event narratives accumulated over time. Due to the massive amount and chaotic nature of the text data, it is difficult to deduce recurrent recurrence trends from it. Using the pipeline industry as an example, Liu et al. [10] used NLP and textual mining algorithms to take ad-vantage of the resource and comprehend the occurrences' underlying causes and contributing aspects. The Pipeline and Hazardous Materials Safety Administration's (PHMSA) incident database's 3587 event narrative entries in the "comment" section were used for this. It has been shown from prior research articles that only a small number of studies have used NLP in the field of material science to achieve knowledge summarization goals. Therefore, this is the first research study to use NLP for summarizing knowledge on High Entropy alloys [11-30]. The future scope of this work is to implement this approach in other domains of material science for knowledge extraction purposes [31-36].

## 2. High Entropy Alloys and its applications

In a multi-component system, high entropy alloys (HEAs) are next generation alloys that contain multiple primary elements. MPEAs (multi-principal element alloys) and complex concentrated alloys are other names for HEAs (CCAs). Professor Ye Junwei first put forth the idea of high entropy alloys (HEAs) and multi-principal element alloys (MPEAs) in 2004 [37-38]. Since each HEA can be adjusted by small elemental additions, like with present element-based alloys, each HEA is a new alloy base. Numerous novel alloy bases are provided by HEAs [39]. Equation 1 provides the number of HEA systems (unique combinations of elements without disclosing composition).

$$C\binom{n}{r} = \frac{n!}{r!(n-r)!} \tag{1}$$

where n is the number of components in the array that are being used to choose the r primary alloy elements. In order to create a conventional alloy, researchers participating in alloy creation historically concentrated on the phase diagram's corners, which only take up a small part of the design space (see figure 1). The recent advancement of HEAs, however, has caused the center area to receive more attention [40].

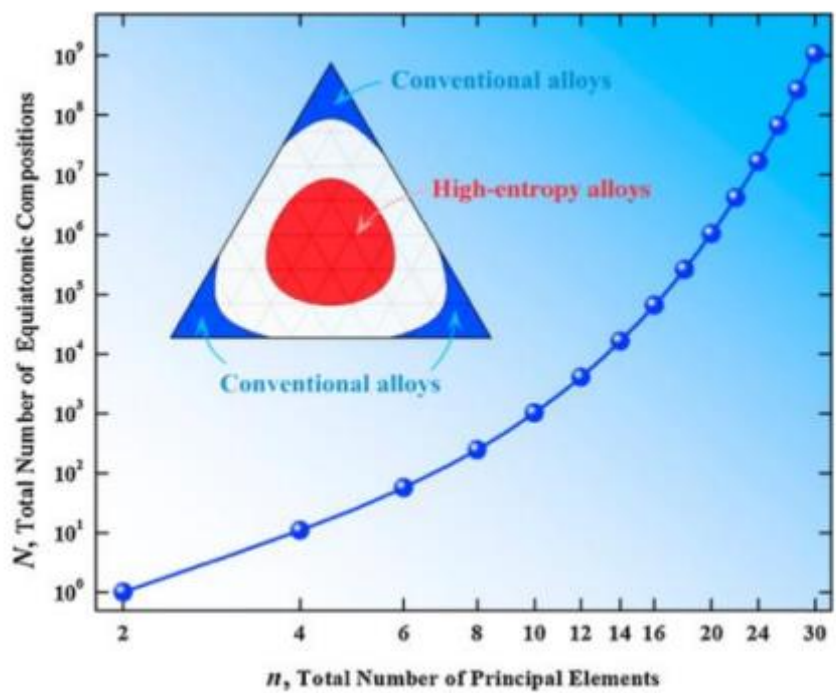

**Figure 1.** The relationship between the total number of primary elements and the total number of equiatomic compositions. The ternary figure in the inset shows how the design of high-entropy alloys differs from that of normal alloys [40].

Every one of the at least five metallic materials used to create high entropy alloys (HEAs) has an atomic concentration that ranges from 5 to 35% in the molar form. High entropy alloys have great work hardenability, high temperature oxidation resistance, and good ductility. Additionally, HEAs have outstanding attractive magnetic characteristics, a high level of wear resistance, and superior erosion resistance. The three main categories of HEA fabrication techniques are liquid mixing, solid mixing, and gaseous mixing. Arc melting, electric resistance melting, inductive melting, Bridgman solidification, and laser additive manufacturing are all types of liquid mixing [41–46].

As indicated in figure 2, there are four categories of entropic alloys: high entropy al-loys, medium entropy alloys, low entropy alloys, and pure metal. When it comes to low temperature applications, high entropy alloys perform better than conventional materials.

Cryogenic processes use alloys with high entropy [47]. They can be employed as low temperature materials in the civil engineering, superconducting, and aircraft sectors. For a variety of military applications, Geanta et al. [48] tested and characterized High-Entropy Alloys from the AlCrFeCoNi System. According to a microstructure research, the microstructure of alloys with high entropy appeared frozen when they were in a molten state as illustrated in figure 3. As illustrated in figure 4, impact testing results revealed that high entropy alloy is the best choice for ballistic packages robust at high velocity penetration impacts.

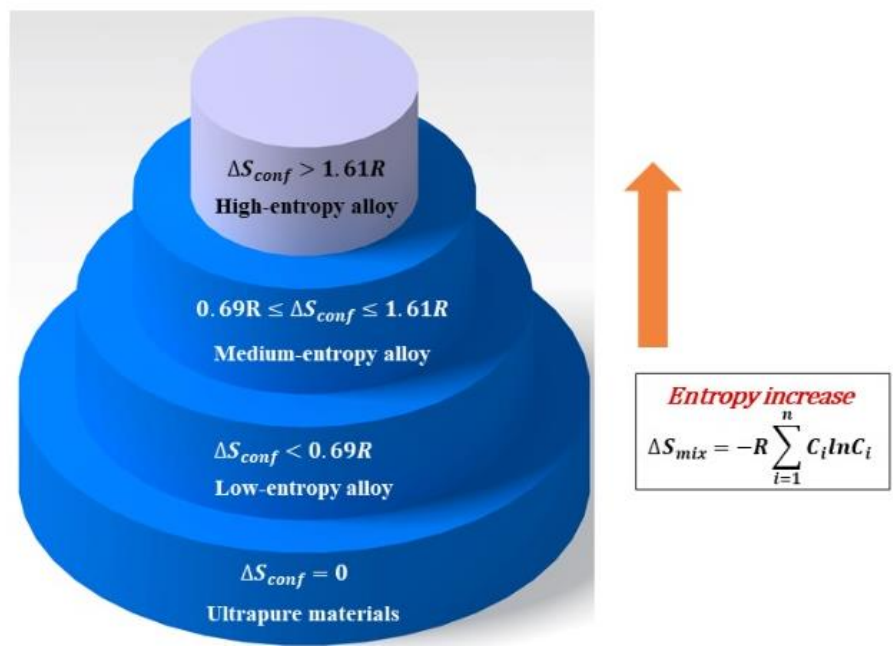

**Figure 2.** Types of High Entropy Alloys [47]

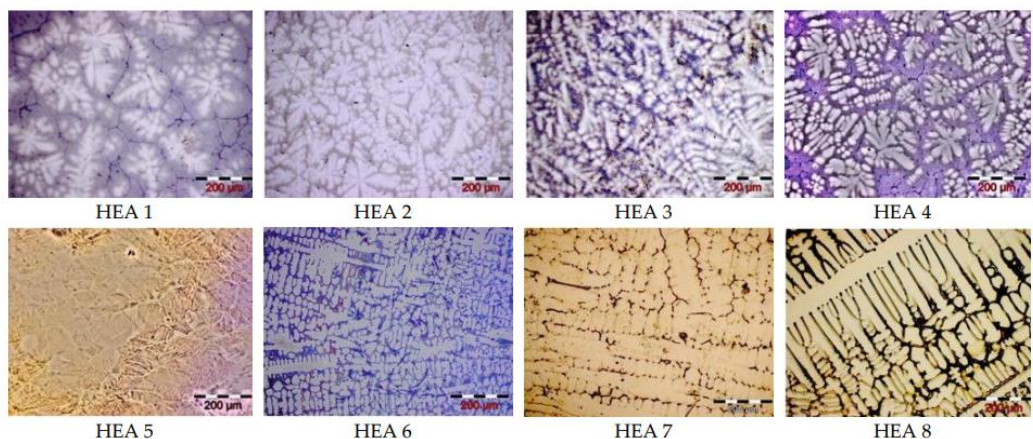

**Figure 3.** Microstructure analysis of an experimental as-cast AlCrxFeCoNi alloys [48]

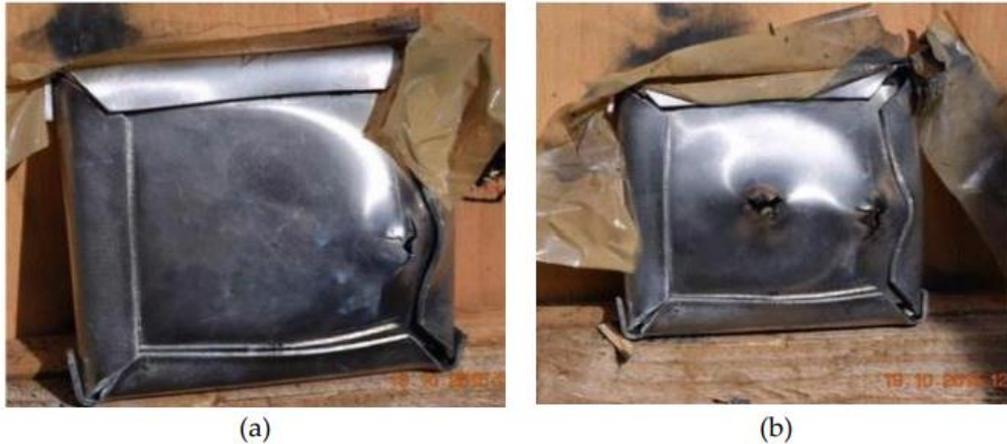

**Figure 4.** Dynamic testing on a HEA-steel ballistic package (a) after the first fire and (b) just after second fire [48]

## 3. Understanding Natural Language Processing Algorithms

Computer science's field of natural language processing (NLP) investigates how computers and human languages interact. It is the engine that powers search engines like Google. Manual linguistic analysis is possible and has been practiced for centuries. But technological advancements continue, particularly in the field of natural language pro-cessing (NLP). Natural Language Processing (NLP) has been aggressively pursued by the machine and deep learning fields using a variety of methods. Even though some of the methods we use today have only been around for a short while, they are already altering how we communicate with technology. Research in the area of natural language pro-cessing (NLP) shows us how to create systems that can comprehend human language in practical ways. Among many others, these include chatbots, machine translation soft-ware, and speech recognition systems. The same problem, however, can be solved using a wide variety of algorithms. In this article, four widely used techniques for teaching ma-chine learning models to analyze data from human language are compared.

NLP (natural language processing) algorithms come in several forms. They can be divided into groups based on the tasks they do, such as entity recognition, relation extraction, entity parsing, and part of speech tagging. The tags created by Part of Speech Tagging algorithms describe how specific phrase components work. For instance, the usual tags for words in English phrases are verb, noun, and preposition. The process of parsing is breaking down a sentence into its grammatical constituents (phrase structure), typically by splitting it into phrases and pieces of speech that trees can represent. Named entities, such as persons, companies, or locations, are found within a text using named entity recognition. Semantic analysis's subsidiary task is relationship extraction. According to relation extraction, there must be a connection between two items if they may be connected by an auxiliary verb (such as "is," "was," or "was not," etc.) or another marker word (such as "to" or "from").

Several NLP algorithms are created with a variety of goals in mind, from language production to sentiment analysis. One technique for helping computers comprehend your intentions when you speak or write is sentiment analysis. Companies employ sentiment analysis as a tool to ascertain whether their customers are satisfied with the goods or ser-vices they provide. However, it also has the potential to better understand how individu-als feel about issues in politics, the medical field, or any other field where individuals have different convictions.

Based on their language, as well as their social media postings and comments, sentiment analysis may tell whether someone is feeling positively or neg-atively about anything. Sentiment analysis is important because it enables businesses to handle dissatisfied clients more effectively. Named entity recognition is sometimes thought of as text classification, where a group of documents must be categorized accord-ing to person or organization names. Algorithms for machine learning typically handle this task. There are many classifiers available, but the k-nearest neighbor technique is the most straightforward (kNN). This straightforward technique is predicated on the idea that numerous words adjacent to a specific thing described in a document are also parts of that entity. Text processing tasks like text summarization have received a lot of attention recently. The fundamental goal of text summarizing is to provide an edited version of the initial text that only conveys the key ideas. The two types of text summarization are extractive and abstractive. Term frequency-inverse document frequency (TF-IDF) score, frequency words, and other statistical measures linked to centrality (i.e., how significant they are) are used to choose sentences from the source documents in the extract text summarization method, also known as key phrase extraction. Abstractive text summarization produces new phrases with details not found in the source text. A subset of natural language pro-cessing is called aspect mining. Aspect mining identifies the various traits, components, or facets of text. Aspect mining divides texts into various categories in order to pinpoint the attitudes—often referred to as sentiments—described in each category. When aspects and subjects are compared, the topic is classified rather than the feeling. Aspects might include things, acts, feelings or emotions, qualities, occurrences, and more, depending on the technique employed. Companies have used aspect mining methods to identify customer responses. Aspects can be important components in the research and development of artificial intelligence for new technologies like chatbots, which use AI programs to give a response with appropriate answers based on information gleaned from text conversations that they analysis for patterns and interconnection between questions and answers. To obtain explicit or implicit attitudes regarding features of text, aspect mining is some-times integrated with sentiment analysis tools, a different sort of natural language pro-cessing. Because aspects and views are so closely related, they are frequently employed as synonyms in literature. Companies might benefit from aspect mining since it enables them to identify the type of client replies.

## 4. Metrics features for evaluating the performance of knowledge summarization in the present study

In the present work, we have implemented metrics features such as BLEU and ROUGE for measuring the performance of NLP algorithms for knowledge summarization purposes. Both the ROGUE and BLEU sets of metrics can be used to create text summaries. BLEU was first required for machine translation, but it is as suitable for the task of text summarization.

A candidate translation of a text is scored using the Bilingual Evaluation Understudy (BLEU), which compares it to one or even more reference translations. Despite being designed for translation, it has the potential to assess text produced for a variety of NLP activities. The computation of Precision scores for the 1- to 4-gram weight range stated in equations 2 is the initial stage.

$$Precision\ of\ N-Grams\ (p_n) = \frac{Number\ of\ correct\ predicted\ n-grams}{Number\ of\ total\ predicted\ n-grams} \quad (2)$$

Next, we use equation 3 to aggregate these Precision Scores. For various N values and with various weight values, this can be calculated.

$$Geometric\ Average\ Precision\ (N) = exp(\sum_{n=1}^{N} w_n log p_n) \qquad (3)$$

Equation 4 is used in the third step to calculate a "Brevity Penalty." With an exponential decline, the shortness penalty penalizes computed translations which are too brief in comparison to the nearest reference length. The shortness penalty makes up for the absence of a recall phrase in the BLEU score.

$$Brevity\ Penalty = \begin{cases} 1, & if\ c > r \\ e^{\left(1-\frac{r}{c}\right)}, & if\ c \leq r \end{cases} \qquad (4)$$

Where c is the predicted length and r is the target length.

Then, as illustrated in equation 5, we multiply the Brevity Penalty by the Geometric Average of the Precision Scores to arrive at the Bleu Score.

$$BLEU(N) = Brevity\ Penalty \cdot Geometric\ Average\ Precision\ (N) \qquad (5)$$

Another metric feature used is the ROUGE algorithm. Recall-Oriented Understudy for Gisting Evaluation is the abbreviation for this method. In essence, it is a collection of measures for assessing machine translations and text summarizations that are done automatically. It functions by contrasting an automatically generated summary or translation with a collection of reference summaries (typically human-produced). The number of "n-grams" that match between the text produced by our model and a "reference" is measured by ROUGE-N. A collection of tokens or words is known as an n-gram. A bigram (2-gram) consists of two consecutive words, whereas a unigram (1-gram) only has one word.

## 5. Materials and Methods

Creating a concise, fluid, and, most importantly, excellent description of lengthy text content is known as text summarization. The fundamental goal of information extraction is to be capable of extracting the most important information from a large body of text and displaying it in a human-readable way. Automatic text summarizing techniques could be particularly beneficial as online textual documents increase since more informative material can be viewed quickly. The framework of the knowledge summarization by NLP algorithms is shown in Figure 5. Data were collected from the abstract of the twenty published papers based on High Entropy Alloys available on the Google Scholar database which has wide availability of papers in comparison to other databases. Preprocessing of text has a significant impact on how well models perform. In order to develop a machine learning model, preprocessing the data is a crucial stage, and the effectiveness of the preprocessing determines the final outcomes. The initial stage of creating a model in NLP is text preparation. In any language used by people, stop words are common. These words are eliminated from our text, allowing us to concentrate more on the crucial information by re-moving the low-level information they contain. The most significant of these is tokenization. It is the process of segmenting a stream of text information into tokens, which can be words, keywords, sentences, symbols, or other significant components. Numerous open-source programs can be used to tokenize data. Text is divided into sentences through the process of sentence tokenization.

Because the tokenizer is based on a corpus of professional English text, it performs well when applied to literature, reporting, and formal documents. Tokenization is the procedure of dividing text into a sequence of tokens from a string of text. Tokens can be viewed as components, similar to how a word functions as a token in a phrase and how a sentence functions as a token in a prose. For evaluating the performance, BLEU and ROUGE metrics features are then implemented. The accuracy of BLEU is slightly in-creased by using the BLEU-4 feature. Recall, precision, F1 score are estimated using equations 6, 7, and 8, respectively, and further evaluation of the ROUGE Score.

$$Recall\ Value = \frac{number\ of\ n-grams\ found\ in\ model\ and\ references}{number\ of\ n-grams\ in\ references} \quad (6)$$

$$Precision\ Value = \frac{number\ of\ n-grams\ found\ in\ model\ and\ references}{number\ of\ n-grams\ in\ model} \quad (7)$$

$$F1 - Score = 2 \times \frac{Precision\ Value \times Recall\ Value}{Precision\ Value + Recall\ Value} \quad (8)$$

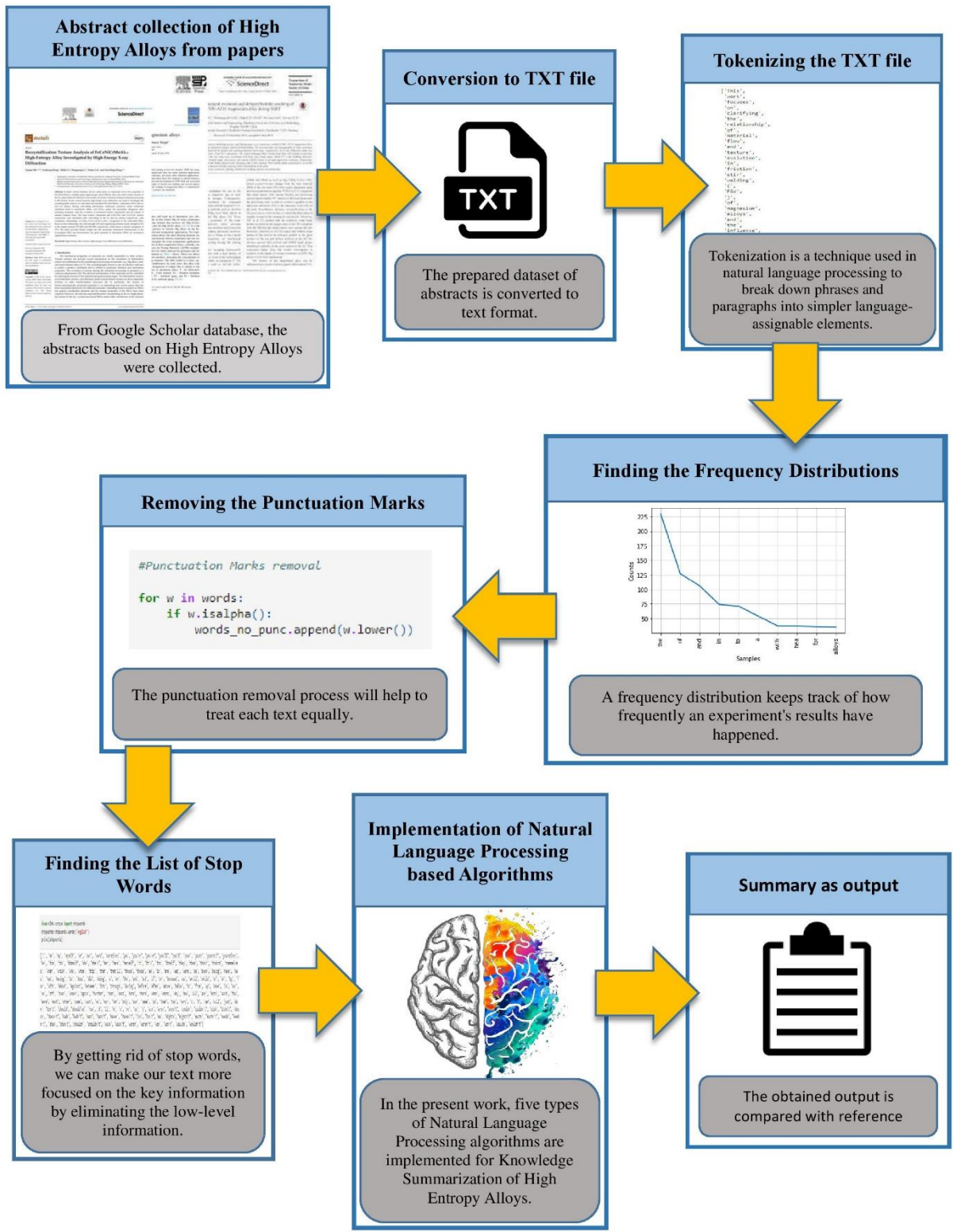

**Figure 5.** NLP framework used in the recent study

## 6. Results

Figure 6 presents the initial plot of the frequency distribution. We can observe that the distribution includes a lot of punctuation and non-content terms like "the," "of," "and," etc. (we refer to them as stop words). Before drawing the graph, we can take them out. To do this, we must import stopwords out from corpus package. A new list of things to ignore is created by adding the stop word list, a list of punctuation, and a list of single figures uti-lizing + signs. Figure 7 displays the final plot of the frequency distribution after punctua-tion and stop marks have been removed. Every human language contains a large number of stop words. By eliminating the low-level information, these phrases can be modified to be more concentrated on the important information.

The filtered tokens can alternatively be displayed as a word cloud. This enables us to use the WordCloud().generate from frequencies() method to get a general overview of the corpus. A frequency dictionary containing all the tokens and their frequencies in the text serves as the method's input. It is necessary to first import the Counter package into Py-thon before generating a vocabulary using the input of the filtered text variable. After pre-processing the word count plot is obtained as seen in figure 8. Word in a word cloud is used to visualize text which shows the frequency of that word. To highlight important textual information, use a word cloud.

Figure 9 shows the plot of the obtained Individual N- grams values for the corre-sponding algorithms. It is observed that the individual 1-gram value of the LUHN algorithm is the highest in comparison to the other algorithms. Figure 10 shows the ob-tained BLEU score for each algorithm. To slightly increase the performance of the BLEU score, BLEU-4 score is obtained as shown in Figure 11.

Table 1 displays the results of the summative evaluation of the Natural Language Processing algorithms utilized for information summarization using the ROGUE method.

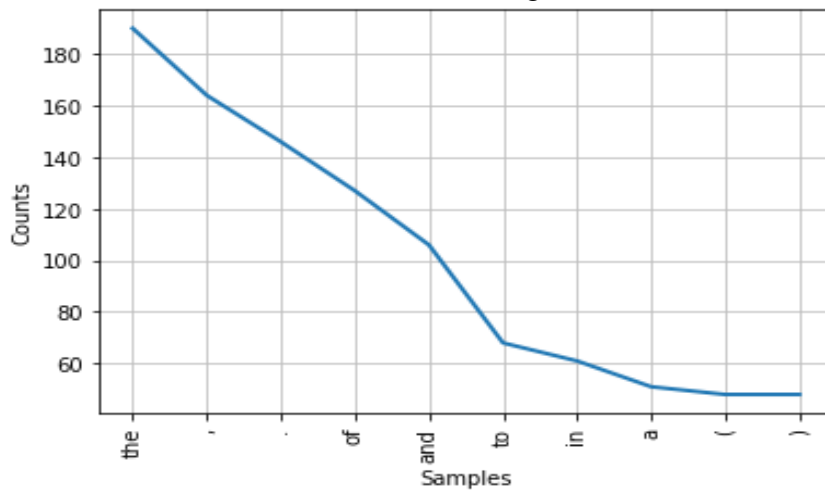

**Figure 6.** Initial Frequency Plot

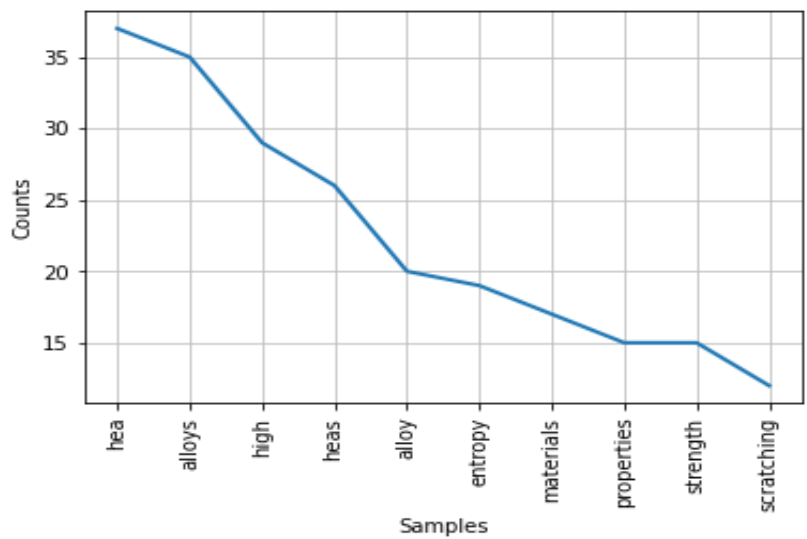

**Figure 7.** Final Frequency Plot

**Figure 8.** Word Cloud

**Figure 9.** Obtained N-grams for each NLP algorithms

**Figure 10.** BLEU Score for the implemented NLP Algorithms

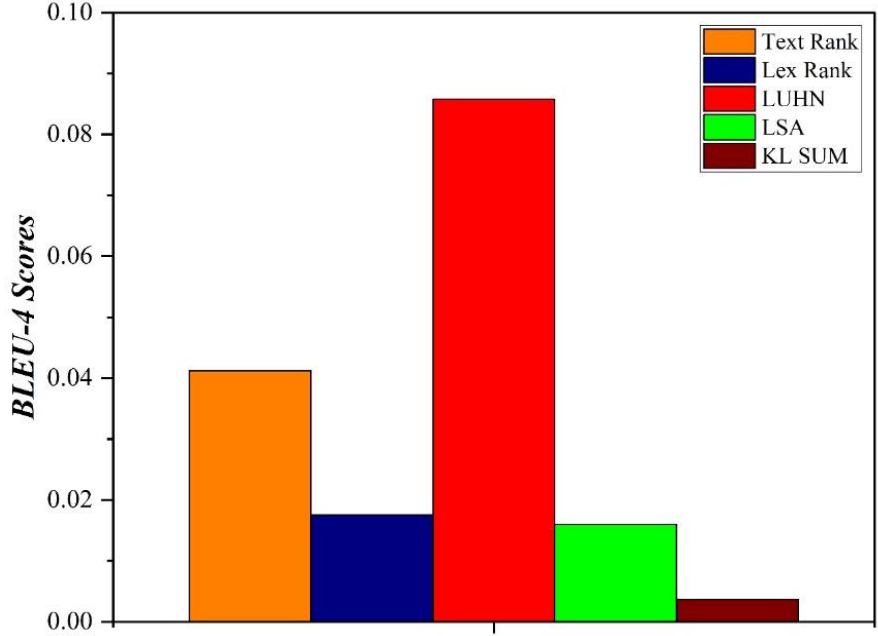
**Figure 11.** BLEU-4 Scores for the implemented NLP Algorithms

**Table 1**. NLP implementation results

| NLP Algorithms | Recall | Precision | F1-Score |
|---|---|---|---|
| Text Rank | 1.0 | 0.370 | 0.540 |
| Lex Rank | 1.0 | 0.299 | 0.460 |
| LUHN | 1.0 | 0.423 | 0.595 |
| Latent Semantic Analysis | 1.0 | 0.350 | 0.519 |
| KL Algorithm | 1.0 | 0.250 | 0.400 |

## 7. Discussion

It is possible to find both keywords and most relevant phrases within a text using a text ranking algorithm based on graphs. It is used to identify the most important sentences in a manuscript by creating a graph with each sentence's vertices being the network's nodes, and the edges connecting them being the number of words they share.

With the help of the Pagerank algorithm, the most important sentences are selected based on their phrase structure. Now, we are only allowed to use the most important sentences while writing a summary. In the Page Rank algorithm, weights are assigned in accordance with Equation 9.

$$X(V_i) = (1 - d) + d \times \sum_{j \in In(V_i)} \frac{1}{|Out(V_j)|} X(V_j) \tag{9}$$

Where X(V_i) is the assigned weight of the i webpage, d is the damping factor, In(V_i) is the inbound links of i, and Out(V_j) is the outgoing links of j. In order to find pertinent terms, the textrank algorithm creates a word network. This network is created by looking at the sequence in which words come before one another. In the text, a connection is made between two words when they are placed next to one another. The connection is more significant if the two terms are placed next to one another more frequently.

In order to determine the significance of each word, the Pagerank method is applied to the resulting network. Despite their relevance, they are all retained except the top third. As soon as they appear consecutively in the text, the relevant terms are arranged into a keywords table. Now, we are only permitted to include key sentences when summarizing the content. In order to find pertinent terms, the textrank algorithm creates a word network. This network is constructed by looking at the order in which words come before one another. In the text, a connection is made between two words when they are placed next to one another. If the two terms are placed next to one another more frequently, the connection is given more significance. The relevance of each word in the resulting network is then determined using the Pagerank algorithm. All but the top third of these words are kept since they are thought to be relevant. If the pertinent words appear right after one another in the text, they are then combined into a keywords table.

The LexRank algorithm assesses the centrality of sentences on graphs as a tool for unsupervised text summarization. The fundamental idea is that sentences "endorse" additional sentences similar to them for the reader. Because of this, a statement is likely to be significant if it sounds like many others. The unsupervised LexRank technique uses eigenvector centrality-based images to intelligently summarize the input text. A cosine similarity-based adjacency matrix is used to visualize the six required sentences. The sentences are ranked according to how similar they are to the centroid sentence, which serves as the mean for all the assertions in the text file in this method. The N (number of all possible words)-dimensional vectors are represented by the bag of words model, which is used to define similarity. Equation 10 is used to determine the value of the appropriate dimension in the text's vector representation.

$$Values\ of\ Corresponding\ Dimension = number\ of\ occurrences\ of\ the\ word \times Inverse\ Document\ Frequency\ (IDF) of\ the\ word \qquad (10)$$

The associated IDF-modified cosine number of the word is determined by Equation 11.

$$IDF\ Modified\ Cosine\ Value\ (x,y) = \frac{\sum_{w \in x,y} tf_{w,x} tf_{w,y} (idf_w)^2}{\sqrt{\sum_{x_i \in x}(tf_{x_i,x} idf_{x_i})^2} \sqrt{\sum_{y_i \in y}(tf_{y_i,y} idf_{y_i})^2}} \qquad (11)$$

where $idf_w$ is the inverse document frequency of a word w and $tf_w$,s is the number of times the word w appears in the sentences.

By analyzing sample corpora of natural literature, latent semantic analysis (LSA) is a computational method for computer simulation and modeling of the definition of words and passages. Many facets of learning and interpreting human languages can be accurately approximated by LSA. It supports a range of applications in information retrieval, instructional technology, and other information processing challenges where complicated wholes can be viewed as additive functionalities of constituent elements. Latent Semantic Analysis, commonly known as LSI (for Latent Semantic Indexing), simulates the contribution of words combined into coherent sequences to natural language. It uses the well-known Singular Value Decomposition (SVD) method from matrix algebra, which wasn't until the late 1980s, and the development of potent digital computing systems and algorithms to make use of them that it was possible to apply to such complicated events. To create a semantic space for a speech, LSA divides a sizable representational text corpus into rectangular cells, each containing a transform of the frequency of a given word in a specific section.

The Luhn algorithm uses a TF-IDF-based methodology. According to their frequency, it only chooses words with tremendous significance. The words appearing at the document's beginning are given more weight. The initial step is to identify which words con-tribute most to the document's meaning. A frequency analysis must be conducted first, followed by the placement of significant but not insignificant English terms.

In the second step, the terms that appear most often in the document are determined, and a subset of those that are less common but still important is then chosen. Calculating a sentence's grade requires the usage of Equation 12.

$$Score = \frac{(Number\ of\ meaningful\ words)^2}{Span\ of\ meaningful\ words} \qquad (12)$$

The Luhn approach makes creating summaries out of a group of words simple. The technique can be applied in two steps. First, we attempt to determine which keywords are most crucial to understanding the text as a whole. In order to accomplish this, according to Luhn, a frequency study should be done first, then phrases that contain major but not inconsequential English terms must be found. A subset of the less frequent but still signif-icant terms is then selected after the most frequent terms in the material are identified in the second step.

Every time KL divergence decreases, a sentence is added to the summary using the Kull-back-Lieber (KL) method, which would be predicated on the greedy method strategy. The KL approach decides whether or not it detracts from its own intake vocabulary by re-ducing the summary vocabulary. If there are L elements in a synthesized and D for a par-ticular text, the KL algorithm offers a criterion for selecting summaries. Although various error metrics are in use right now, sending as little information as possible is our top pri-ority. These two models reduce our problem to a finite number of variables. The best method for deciding which distribution is better is to use the Kullback-Liber algorithm to find which distribution maintains the most information from the original data source.

The findings imply that phrase centrality rating derived from graphs are used by the LexRank algorithm for unsupervised text summarization. The fundamental tenet is that statements "recommend" other assertions to readers if they are comparable to their own. A remark is likely extremely important if it is virtually identical to multiple others. The significance of the word "recommending" in this statement helps to illustrate its signifi-cance. As a result, many other phrases must be linked to a statement for it to communicate and be included with the summary. This makes perfect sense and allows applying the technique to any brand-new content.

According to reports, the LexRank algorithm's f1-score and accuracy value are lower than that of the LSA approach. To understand textual information statistically, Latent Semantic Analysis (LSA) is used. The foundation of topic modeling is this tactic. A specific topic and data matrix of our current arrangements and materials will be split into two separate matrices to achieve the main purpose.

The Luhn Algorithm has higher f1-scores than the LexRank and LSA algorithms, it has been found. The Luhn technique's Term Frequency-Inverse Document Frequency (TF-IDF) basis may also be shown. Only the most important terms are selected, based on how consistently they are used. A phrase's first few words are given additional weight. The Text rank method likewise follows the f1-score of the Luhn algorithm. It is a fairly evident objective for Text

Rank to identify how closely related each expression is to each other phrase in the text. It is observed that in addition to the ROUGE score also the BLEU-4 score of the LUHN algorithm is higher in comparison to the other algorithms.

## 8. Conclusions

An autonomous text summary can be produced with the help of machine learning and natural language processing. Processing natural language is crucial to ma-chine-human interaction. It is still developing, and more research is being done in this ar-ea. Natural Language Processing and automatic knowledge summary have emerged as vital methods for compressing lengthy and tedious papers, whether technical, financial, legal, medical or literary. Every sector, including business and academia, stands to win. And we are still far from realizing all of its possibilities. In the future, text summarization technology may become more advanced and capable of comprehending natural language. The most recent research leads to the following conclusions:

- •The research publications that summarized High Entropy Alloys were sub-jected to Text Rank, KL Algorithm, LSA, Luhn, and Lex Rank i.e. five distinct Natural Language Processing techniques.

- •The results showed that the Luhn algorithm produced the highest f1-score of 0.595 compared to the other algorithms.

- •The BLEU-4 results support the results obtained by ROUGE metric features. It is observed that the BLEU-4 score of the LUHN algorithm is higher in comparison to the other algorithms.

- •The main limitation is the dependency of accuracy score on the number of datasets. This limitation can be improved by increasing the number of datasets.

- •The future scope of this work will be the implementation of more sophisticated algorithms like BERT, XL Net and GPT2 for the knowledge summarization purpose.

**Funding:** This research received no external funding

**Data and Code Availability Statement:** The data and code supporting this study's findings are available upon request from the authors.

**Conflicts of Interest:** The authors declare no conflict of interest.

## References

1. Chowdhary K. Natural language processing. Fundamentals of artificial intelligence. 2020:603-49.
2. Nadkarni PM, Ohno-Machado L, Chapman WW. Natural language processing: an introduction. Journal of the American Medical Informatics Association. 2011 Sep 1;18(5):544-51.
3. Chopra, A., Prashar, A. and Sain, C., 2013. Natural language processing. International journal of technology enhancements and emerging engineering research, 1(4), pp.131-134.

4. Bird, S., Klein, E. and Loper, E., 2009. Natural language processing with Python: analyzing text with the natural language toolkit. "O'Reilly Media, Inc.".
5. Lehnert, W.G. and Ringle, M.H. eds., 2014. Strategies for natural language processing. Psychology Press.
6. Olivetti, E.A., Cole, J.M., Kim, E., Kononova, O., Ceder, G., Han, T.Y.J. and Hiszpanski, A.M., 2020. Data-driven materials research enabled by natural language processing and information extraction. Applied Physics Reviews, 7(4), p.041317.
7. Xie, L., Zhu, X., Sun, W., Jiang, C., Wang, P., Yang, S., Fan, Y. and Song, Y., 2022. Investigations on the material flow and the influence of the resulting texture on the tensile properties of dissimilar friction stir welded ZK60/Mg–Al–Sn–Zn joints. Journal of Materials Research and Technology, 17, pp.1716-1730.
8. Mohammadi, E., 2012. Knowledge mapping of the Iranian nanoscience and technology: a text mining approach. Scientometrics, 92(3), pp.593-608.
9. Maghrebi, M., Waller, S.T. and Sammut, C., 2015. Text mining approach for reviewing the ready mixed concrete literature. In 2nd International Conference on Civil and Building Engineering Informatics, University of Osaka, Tokyo, Japan (pp. 105-109).
10. Liu, G., Boyd, M., Yu, M., Halim, S.Z. and Quddus, N., 2021. Identifying causality and contributory factors of pipeline incidents by employing natural language processing and text mining techniques. Process Safety and Environmental Protection, 152, pp.37-46.
11. Sathiyamoorthi, P. and Kim, H.S., 2022. High-entropy alloys with heterogeneous microstructure: processing and mechanical properties. Progress in Materials Science, 123, p.100709.
12. Zhang, Y., Wang, D. and Wang, S., 2022. High-Entropy Alloys for Electrocatalysis: Design, Characterization, and Applications. Small, 18(7), p.2104339.
13. Dewangan, S.K., Mangish, A., Kumar, S., Sharma, A., Ahn, B. and Kumar, V., 2022. A review on high-temperature applicability: a milestone for high entropy alloys. Engineering Science and Technology, an International Journal, p.101211.
14. Zhang, Q., Zhang, S., Luo, Y., Liu, Q., Luo, J., Chu, P.K. and Liu, X., 2022. Preparation of high entropy alloys and application to catalytical water electrolysis. APL Materials, 10(7), p.070701.
15. Xue, L., Ding, Y., Pradeep, K.G., Case, R., Castaneda, H. and Paredes, M., 2022. Development of a non-equimolar AlCrCuFeNi high-entropy alloy and its corrosive response to marine environment under different temperatures and chloride concentrations. Journal of Alloys and Compounds, p.167112.
16. Wu, Q., He, F., Li, J., Kim, H.S., Wang, Z. and Wang, J., 2022. Phase-selective recrystallization makes eutectic high-entropy alloys ultra-ductile. Nature communications, 13(1), pp.1-8.
17. Oliveira, J.P., Shen, J., Zeng, Z., Park, J.M., Choi, Y.T., Schell, N., Maawad, E., Zhou, N. and Kim, H.S., 2022. Dissimilar laser welding of a CoCrFeMnNi high entropy alloy to 316 stainless steel. Scripta Materialia, 206, p.114219.
18. Wu, D., Kusada, K., Nanba, Y., Koyama, M., Yamamoto, T., Toriyama, T., Matsumura, S., Seo, O., Gueye, I., Kim, J. and Rosantha Kumara, L.S., 2022. Noble-Metal High-Entropy-Alloy Nanoparticles: Atomic-Level Insight into the Electronic Structure. Journal of the American Chemical Society, 144(8), pp.3365-3369.
19. Rong, Z., Wang, C., Wang, Y., Dong, M., You, Y., Wang, J., Liu, H., Liu, J., Wang, Y. and Zhu, Z., 2022. Microstructure and properties of FeCoNiCrX (X Mn, Al) high-entropy alloy coatings. Journal of Alloys and Compounds, 921, p.166061.
20. Wei, D., Gong, W., Wang, L., Tang, B., Kawasaki, T., Harjo, S. and Kato, H., 2022. Strengthening of high-entropy alloys via modulation of cryo-pre-straining-induced defects. Journal of Materials Science & Technology, 129, pp.251-260.

21. Yang, Y., Li, Z., Zhang, W., Ma, Y., Xiong, Q., Li, W., Chen, S. and Wu, Z., 2022. Concentration of "Mysterious Solute" in CoCrFeNi high entropy alloy. Scripta Materialia, 211, p.114504.
22. Jin, Z., Zhou, X., Hu, Y., Tang, X., Hu, K., Reddy, K.M., Lin, X. and Qiu, H.J., 2022. A fourteen-component high-entropy alloy@ oxide bifunctional electrocatalyst with a record-low Δ E of 0.61 V for highly reversible Zn–air batteries. Chemical Science.
23. Banko, L., Krysiak, O.A., Pedersen, J.K., Xiao, B., Savan, A., Löffler, T., Baha, S., Rossmeisl, J., Schuhmann, W. and Ludwig, A., 2022. Unravelling Composition–Activity–Stability Trends in High Entropy Alloy Electrocatalysts by Using a Data-Guided Combinatorial Synthesis Strategy and Computational Modeling. Advanced Energy Materials, 12(8), p.2103312.
24. Wetzel, A., von der Au, M., Dietrich, P.M., Radnik, J., Ozcan, O. and Witt, J., 2022. The comparison of the corrosion behavior of the CrCoNi medium entropy alloy and CrMnFeCoNi high entropy alloy. Applied Surface Science, 601, p.154171.
25. Zhou, X.Y., Zhu, J.H., Wu, Y., Yang, X.S., Lookman, T. and Wu, H.H., 2022. Machine learning assisted design of FeCoNiCrMn high-entropy alloys with ultra-low hydrogen diffusion coefficients. Acta Materialia, 224, p.117535.
26. Moghaddam, A.O., Sudarikov, M., Shaburova, N., Zherebtsov, D., Zhivulin, V., Solizoda, I.A., Starikov, A., Veselkov, S., Samoilova, O. and Trofimov, E., 2022. High temperature oxidation resistance of W-containing high entropy alloys. Journal of Alloys and Compounds, 897, p.162733.
27. Wei, D., Wang, L., Zhang, Y., Gong, W., Tsuru, T., Lobzenko, I., Jiang, J., Harjo, S., Kawasaki, T., Bae, J.W. and Lu, W., 2022. Metalloid substitution elevates simultaneously the strength and ductility of face-centered-cubic high-entropy alloys. Acta Materialia, 225, p.117571.
28. Yuan, J.L., Wang, Z., Jin, X., Han, P.D. and Qiao, J.W., 2022. Ultra-high strength assisted by nano-precipitates in a heterostructural high-entropy alloy. Journal of Alloys and Compounds, 921, p.166106.
29. Hou, J.X., Liu, S.F., Cao, B.X., Luan, J.H., Zhao, Y.L., Chen, Z., Zhang, Q., Liu, X.J., Liu, C.T., Kai, J.J. and Yang, T., 2022. Designing nanoparticles-strengthened high-entropy alloys with simultaneously enhanced strength-ductility synergy at both room and elevated temperatures. Acta Materialia, 238, p.118216.
30. Doan, D.Q. and Fang, T.H., 2022. Effect of vibration parameters on the material removal characteristics of high-entropy alloy in scratching. International Journal of Mechanical Sciences, 232, p.107597.
31. Dhungana, D. S., Mallet, N., Fazzini, P.-F., Larrieu, G., Cristiano, F., & Plissard, S. (2022). Self-catalyzed InAs nanowires grown on Si: the key role of kinetics on their morphology. Nanotechnology. https://doi.org/10.1088/1361-6528/ac8bdb
32. Sadek, D., Dhungana, D. S., Coratger, R., Durand, C., Proietti, A., Gravelier, Q., Reig, B., Daran, E., Fazzini, P. F., & Cristiano, F. (2021). Integration of the rhombohedral BiSb (0001) Topological Insulator on a cubic GaAs (001) substrate. ACS Applied Materials & Interfaces, 13(30), 36492–36498.
33. Dhungana, D. S., Grazianetti, C., Martella, C., Achilli, S., Fratesi, G., & Molle, A. (2021). Two-Dimensional Silicene–Stanene Heterostructures by Epitaxy. Advanced Functional Materials, 31(30), 2102797.
34. Dhungana, D. S., Hemeryck, A., Sartori, N., Fazzini, P.-F., Cristiano, F., & Plissard, S. R. (2019). Insight of surface treatments for CMOS compatibility of InAs nanowires. Nano Research, 12(3), 581–586.
35. Sefene, E., Tsegaw, A., Mishra, A. (2022). 'Process Parameter Optimization of 6061AA Friction Stir Welded Joints Using Supervised Machine Learning Regression-Based Algorithms', Journal of Soft Computing in Civil Engineering, 6(1), pp. 127-137. doi: 10.22115/scce.2022.299913.1350

36. Bonaventura, E., Dhungana, D.S., Martella, C., Grazianetti, C., Macis, S., Lupi, S., Bonera, E. and Molle, A., 2022. Optical and thermal responses of silicene in Xene heterostructures. Nanoscale horizons, 7(8), pp.924-930.
37. Zhang, Yong. (2019). History of High-Entropy Materials. 10.1007/978-981-13-8526-1_1
38. Yeh JW, et al. Nanostructured high-entropy alloys with multiple principal elements: novel alloy design concepts and outcomes. Adv. Eng. Mat. 2004; 6:299–303. doi: 10.1002/adem.200300567.
39. Miracle D. B. (2019). High entropy alloys as a bold step forward in alloy development. Nature communications,10(1),1805. https://doi.org/10.1038/s41467-019-09700-1
40. Ye, Y.F., Wang, Q., Lu, J., Liu, C.T. and Yang, Y., 2016. High-entropy alloy: challenges and prospects. Materials Today, 19(6), pp.349-362.
41. Modupeola Dada, Patricia Popoola, Samson Adeosun and Ntombi Mathe (September 27th 2019). High Entropy Alloys for Aerospace Applications [Online First], IntechOpen, DOI: 10.5772/intechopen.84982. Available from: https://www.intechopen.com/online-first/high-entropy-alloys-for-aerospace-applications
42. Cui, L.; Ma, B.; Feng, S.Q.; Wang, X.L. Microstructure and Mechanical Properties of High-Entropy Alloys CoCrFeNiAl by Welding. Adv. Mater. Res. 2014, 936, 1635–1640.
43. Lippold, J.; Kiser, S.; DuPont, J. Welding Metallurgy and Weldability of Nickel-Base Alloys; Wiley: Hoboken, NJ, USA, 2013.
44. Vendan, S.; Gao, L.; Garg, A.; Kavitha, P.; Dhivyasri, G.; SG, R. Interdisciplinary Treatment to ARC Welding Power Sources; Springer: Singapore, 2018.
45. Kong, X.; Yang, Q.; Li, B.; Rothwell, G.; English, R.; Ren, X. Numerical study of strengths of spot-welded joints of steel. Mater. Des. 2008, 29, 1554–1561
46. Chen, S.; Tong, Y.; Liaw, P. Additive Manufacturing of High-Entropy Alloys: A Review. Entropy 2018, 20, 937.
47. Rui Xuan Li and Yong Zhang (December 3rd 2018). Entropic Alloys for Cryogenic Applications, Stainless Steels and Alloys, Zoia Duriagina, IntechOpen, DOI: 10.5772/intechopen.82351. Available from: https://www.intechopen.com/books/stainless-steels-and-alloys/entropic-alloys-for-cryogenicapplications
48. Victor Geanta and Ionelia Voiculescu (October 23rd 2019). Characterization and Testing of High-Entropy Alloys from AlCrFeCoNi System for Military Applications [Online First], IntechOpen, DOI: 10.5772/intechopen.88622. Available from: https://www.intechopen.com/online-first/characterization-and-testing-of-highentropy-alloys-from-alcrfeconi-system-for-military-applications